\documentclass{llncs}
\usepackage{times}
\usepackage{epic}
\usepackage{graphicx}
\usepackage{latexsym}
\usepackage{amsmath}
\usepackage{verbatim}
\usepackage{xspace}

\begin{document}

\newlength{\halftextwidth}
\setlength{\halftextwidth}{0.47\textwidth}
\def\halffigsize{2.2in}
\def\thirdfigsize{1.5in}
\def\negvspace{0in}
\def\posvspace{0em}

\input epsf





\newcommand{\set}{\mathcal}
\newcommand{\myset}[1]{\ensuremath{\mathcal #1}}

\renewcommand{\theenumii}{\alph{enumii}}
\renewcommand{\theenumiii}{\roman{enumiii}}
\newcommand{\figref}[1]{Figure \ref{#1}}
\newcommand{\tref}[1]{Table \ref{#1}}
\renewcommand{\And}{\wedge}
\newcommand{\myldots}{\ldots}

\newtheorem{mydefinition}{Definition}
\newtheorem{mytheorem}{Theorem}
\newtheorem{mylemma}{Lemma}
\spnewtheorem*{myexample}{Running Example}{\bf}{\it}
\spnewtheorem*{myexample2}{Example}{\bf}{\it}
\newtheorem{mytheorem1}{Theorem}
\newcommand{\myproof}{\noindent {\bf Proof:\ \ }}
\newcommand{\myqed}{\mbox{$\Box$}}

\newcommand{\constraint}[1]{\mbox{\sc #1}}
\newcommand{\alldiff}{\constraint{All-Different}\xspace}
\newcommand{\regular}{\constraint{Regular}\xspace}
\newcommand{\grammar}{\constraint{Cfg}\xspace}
\newcommand{\alldiffandsum}{\constraint{All-Different+Sum}\xspace}
\newcommand{\gcc}{\constraint{GCC}\xspace}
\newcommand{\egcc}{\constraint{eGCC}\xspace}
\newcommand{\costgcc}{\constraint{Cost-GCC}}
\newcommand{\softgcc}{\constraint{SoftGCC}\xspace}
\newcommand{\minmax}{\constraint{MinMax}}
\newcommand{\dom}{\ensuremath{\mbox{dom}}}
\newcommand{\SwitchC}{\constraint{Switch}}
\newcommand{\MaxSwitch}{\constraint{MaxSwitch}}
\newcommand{\MinSwitch}{\constraint{MinSwitch}}
\newcommand{\sumc}{\constraint{Sum}\xspace}
\newcommand{\interdistance}{\constraint{Inter-Distance}}
\newcommand{\nvalue}{\constraint{NValue}\xspace}
\newcommand{\AtMostNValue}{\constraint{AtMostNValue}\xspace}
\newcommand{\AtLeastNValue}{\constraint{AtLeastNValue}\xspace}
\newcommand{\permutation}{\constraint{Permutation}\xspace}
\newcommand{\mypermutation}{\constraint{Same}\xspace}
\newcommand{\sameset}{\constraint{SameSet}\xspace}
\newcommand{\same}{\constraint{Same}\xspace}
\newcommand{\usedby}{\constraint{Used-By}\xspace}
\newcommand{\uses}{\constraint{Uses}\xspace}
\newcommand{\disjoint}{\constraint{Disjoint}\xspace}
\newcommand{\common}{\constraint{Common}\xspace}
\newcommand{\softalldiff}{\constraint{SoftAllDifferent}\xspace}
\newcommand{\notallequal}{\constraint{Not-All-Equal}\xspace}
\newcommand{\softallequal}{\constraint{Soft-All-Equal}}
\newcommand{\gsc}{\constraint{GSC}}
\newcommand{\sequence}{\constraint{Sequence}\xspace}
\newcommand{\precedence}{\constraint{Precedence}\xspace}
\newcommand{\GCC}{\constraint{GCC}\xspace}
\newcommand{\roots}{\constraint{Roots}\xspace}
\newcommand{\range}{\constraint{Range}\xspace}

\newcommand{\tighter}{\mbox{$\preceq$}}
\newcommand{\stighter}{\mbox{$\prec$}}
\newcommand{\incomparable}{\mbox{$\bowtie$}}
\newcommand{\equivalent}{\mbox{$\equiv$}}

\newcommand{\todo}[1]{{\tt (... #1 ...)}}
\newcommand{\myOmit}[1]{}

\title{Decomposition of the \nvalue constraint}

\author{Christian Bessiere\inst{1}
\and
George Katsirelos\inst{2} \and
Nina Narodytska\inst{2}
\and 
Claude-Guy Quimper\inst{3}
\and
Toby Walsh\inst{2}}
\institute{LIRMM, CNRS, Montpellier, email:
bessiere@lirmm.fr
\and NICTA and University of NSW,
Sydney, Australia, email: 
\{george.katsirelos,nina.narodytska,toby.walsh\}@nicta.com.au
\and
\'{E}cole Polytechnique de Montr\'{e}al, email: 
cquimper@gmail.com}


\maketitle
\begin{abstract}
We study decompositions of 
\nvalue, a global
constraint that can be used to model
a wide range of problems where values
need to be counted. 
Whilst decomposition typically 
hinders propagation, we identify one decomposition that
maintains a global view as enforcing bound consistency
on the decomposition achieves bound consistency
on the original global \nvalue constraint. 
Such decompositions
offer the prospect for advanced solving 
techniques like nogood learning and 
impact based branching heuristics. 
They may also help SAT and IP solvers 
take advantage of the propagation of global constraints. 
\end{abstract}

\sloppy
\section{Introduction}

Global constraints are an important feature of 
constraint programming. They capture common patterns
in real world problems, and provide efficient propagators
for pruning the search space. 
Consider, for example, the \nvalue constraint  which counts the number
of values used by a set of variables \cite{pachet1}. 
This global constraint can model problems where values 
represent resources. 
This is a common constraint that can be used to model many
practical problems such as timetabling and frequency allocation. 
Whilst enforcing domain consistency 
on the \nvalue constraint is NP-hard \cite{bhhwaaai2004},
bound consistency is polynomial to achieve. 
At least four different propagation algorithms 
for the \nvalue constraint have been proposed, some
of which achieve bound consistency
\cite{bcp01,bhhkwcpaior2005,bhhkwconstraint2006}.

We have recently proposed simulating propagators
for global constraints with decompositions. 
For instance, we have shown that 
carefully designed decompositions of the 
global \alldiff and \gcc constraints
can efficiently simulated the corresponding 
bound consistency propagators
\cite{bknqwijcai09}. 
We turn now to the \nvalue constraint.
We study a number of different decompositions,
one of which permits the achievement of bound consistency on 
the \nvalue constraint. 
Such decompositions open out a number of promising directions.
For example, they suggest schema for learning nogoods.
As a second example, such decompositions may help construct
nogood and impact based branching heuristics. 
As a third and final example, such decompositions
may permit SAT and IP solvers to take advantage of 
the inferences performed by the propagators of global
constraints. We have, for instance, seen
this with our decompositions of the \alldiff constraint
\cite{bknqwijcai09}.

\section{Background}

We assume values are taken from the set 1 to $d$. 
We write $dom(X_i)$ for the domain of possible values for $X_i$, $min(X_i)$
for the smallest value in $dom(X_i)$,  $max(X_i)$ for the greatest,
and $range(X_i)$ for the interval $[min(X_i),max(X_i)]$.
A \emph{global constraint} is one in which the number of variables $n$
is a parameter. For instance,
the global $\nvalue([X_1,\ldots,X_n],N)$ 
constraint ensures that 
$N=|\{X_i \ | \ 1 \leq i \leq n\}|$ \cite{pachet1}.
Constraint solvers typically use
backtracking search to explore the space
of partial assignments. After each assignment,
propagation algorithms prune the search
space by enforcing local consistency properties like 
domain, range  or bound consistency. A constraint is 
\emph{domain consistent} (\emph{DC})
iff when a variable is assigned any of the values in its domain, there
exist compatible values in the domains of all the other variables of
the constraint. Such an assignment is called
a \emph{support}. 
A constraint is \emph{disentailed} iff 
there is no possible support. 
%
A constraint is \emph{range consistent} (\emph{RC})
iff, when a variable is
assigned any of the values in its domain, there exist compatible
values between the minimum and maximum domain value
for all the other variables of the constraint.
Such an assignment is called
a \emph{bound support}. 
A constraint is \emph{bound consistent} (\emph{BC})
iff   the minimum and maximum value of every variable of the
constraint belong to a bound support.
%
We will compare local consistency properties applied
to sets of constraints, $c_1$ and $c_2$ which are logically equivalent.
As in \cite{debruyne1}, 
a local consistency property $\Phi$ on $c_1$ is as strong as
$\Psi$ on $c_2$ 
iff, given any domains, if $\Phi$ holds on $c_1$ then $\Psi$ holds on $c_2$;
$\Phi$ on $c_1$ is stronger
than $\Psi$ on $c_2$  
iff 
$\Phi$ on $c_1$ is as strong as
$\Psi$ on $c_2$ but not vice versa;
$\Phi$ on $c_1$ is equivalent to
$\Psi$ on $c_2$ 
iff
$\Phi$ on $c_1$ is as strong as
$\Psi$ on $c_2$ and vice versa.
Finally, as constraint solvers usually  enforce local consistency after each
assignment down a branch in the search tree, we will 
compute the total amortised cost of enforcing a
local consistency down an entire branch of the search tree. 
This captures the incremental cost of propagation.  

\section{\nvalue constraint}

Pachet and Roy proposed
the \nvalue constraint (called by them the ``cardinality on
attribute values'' constraint) to model
a combinatorial problem in selecting musical play-lists \cite{pachet1}.
It can also be used to model the number of frequencies used
in a frequency allocation problem or
the number of rooms needed to timetable a set of
exams. It generalizes several other global 
constraints including \alldiff (which ensures
a set of variables take all different values) and \notallequal 
(which ensures a set of variables do not all take the same value).
Enforcing domain consistency on the \nvalue
constraint is NP-hard
(Theorem 3 in \cite{bhhwaaai2004}) even when $N$ is
fixed (Theorem 2 in \cite{bhhkwcpaior2005}). 
In fact, computing the lower bound on $N$ is NP-hard
(Theorem 3 in \cite{bhhwconstraint2007}). 
In addition, enforcing domain consistency on the
\nvalue constraint is not fixed parameter
tractable since it is 
in the $W$[2]-complete complexity class
along with problems like minimum hitting set
(Theorem 2 in \cite{bhhkqwaaai2008}). 
However, a number of polynomial propagation algorithms have
been proposed that achieve bound consistency
and some closely related levels of local
consistency \cite{bcp01,bhhkwcpaior2005,bhhkwconstraint2006}.

\subsection{Simple decomposition}\label{sec:nvalue:simple}

We can decompose the \nvalue constraint 
by introducing 0/1 variables
and posting the following constraints:
\begin{eqnarray}
& X_i=j \rightarrow B_j=1 & \ \ \ \ \forall 1 \leq i \leq n, 1 \leq j \leq d \label{dec1} \\
& B_j=1 \rightarrow \bigvee_{i=1}^n X_i=j  & 
\ \ \ \ \forall 1 \leq j \leq d  \label{dec2}\\
& \sum_{j=1}^m B_j  = N & \label{dec3}
\end{eqnarray}
Unfortunately, this simple decomposition
hinders propagation. 
It can be BC whereas  BC on  the corresponding
\nvalue constraint detects disentailment. 

\begin{mytheorem}
BC on \nvalue is stronger than BC 
on its decomposition into (\ref{dec1}) to (\ref{dec3}).
\end{mytheorem}
\myproof
Clearly BC on \nvalue is at least as strong
as BC on the decomposition. To show strictness,
consider $X_1 \in \{1,2\}$, $X_2 \in \{3,4\}$,
$B_j \in \{0,1\}$ for $1 \leq j \leq 4$, and $N=1$.
Constraints (\ref{dec1}) to (\ref{dec3}) are BC. However, the
corresponding \nvalue constraint has no bound support
and thus enforcing BC on it detects disentailment. 
\myqed

We observe that  enforcing DC instead of BC on constraints
(\ref{dec1}) to (\ref{dec3}) in the example of the proof above  still
does not prune any value.  
To decompose \nvalue 
without hindering propagation, we must look to 
more complex decompositions.

\subsection{Decomposition into \AtMostNValue and \AtLeastNValue}

Our first step in decomposing the \nvalue constraint
is to split it into two
parts: an \AtMostNValue and an \AtLeastNValue constraint.
$\AtLeastNValue([X_1,\ldots,X_n],N)$ holds
iff $N \leq |\{X_i| 1\leq i \leq n\}|$ whilst
$\AtMostNValue([X_1,\ldots,X_n],N)$ holds
iff $|\{X_i| 1\leq i \leq n\}| \leq N$. 

\begin{myexample}
Consider a \nvalue constraint over
the following variables and values:
$$
{\scriptsize
\begin{array}{c|ccccc} 
 & 1 & 2 & 3 & 4 & 5  \\ \hline
X_1 & \ast & \ast & \ast & & \ast  \\ 
X_2 & & \ast & & & \\ 
X_3 & & \ast & \ast & \ast & \\ 
X_4 & & & & \ast & \\
X_5 & & & \ast & \ast & \\
N & \ast & \ast  & & & \ast 
\end{array}
}
$$
Suppose we decompose this into
an \AtMostNValue and an \AtLeastNValue constraint.
Consider the \AtLeastNValue constraint.
The 5 variables can take at most 
4 different values because $X_2,X_3,X_4$, and $X_5$ can only take
values $2, 3$ and $4$. Hence, there is no bound support for
$N=5$. Enforcing BC on the \AtLeastNValue constraint
therefore prunes $N=5$.
Consider now the \AtMostNValue constraint.
Since $X_2$ and $X_4$ guarantee
that we take at least 2 different values,
there is no bound support for $N=1$. 
Hence enforcing BC on an \AtMostNValue constraint
prunes $N=1$.
If $X_1=1$, $3$ or $5$, or $X_5=3$
then any complete assignment uses at least 3 different values. 
Hence there is also no bound support 
for these assignments. Pruning these
values gives bound consistent domains for
the original \nvalue constraint:
$$
{\scriptsize
\begin{array}{c|ccccc} 
 & 1 & 2 & 3 & 4 & 5  \\ \hline
X_1 & & \ast & & &  \\ 
X_2 & & \ast & & & \\ 
X_3 & & \ast & \ast & \ast & \\ 
X_4 & & & & \ast & \\
X_5 & & & & \ast & \\
N & & \ast  & & & 
\end{array}
}
$$
\end{myexample}

To show that decomposing 
the \nvalue constraint into these two 
parts does not hinder propagation in general, 
we will use the following lemma.
Given an assignment $S$ of values, $card(S)$ denotes the number of
distinct values in $S$. Given  a vector of variables  
$X=X_1\ldots X_n$,  
$card_\uparrow(X)=max\{card(S)\mid S\in \Pi_{X_i\in X} range(X_i)\}$ and
$card_\downarrow(X)=min\{card(S)\mid S\in \Pi_{X_i\in X} range(X_i)\}$. 

\begin{mylemma}[adapted from \cite{bhhkwconstraint2006}]
\label{nvalue:prop_1}
Consider $\nvalue([X_1, \ldots, X_n], N)$. If $D(N) \subseteq [card_\downarrow(X), card_\uparrow(X)]$, 
 then  $N$  is BC
.
\end{mylemma}

\myproof
Let $S_{min}$ be an  assignment of $X$ in $\Pi_{X_i\in X} range(X_i)$ 
with  $card(S_{min}) =
card_\downarrow(X)$  and  $S_{max}$ be an assignment of $X$ in
$\Pi_{X_i\in X} range(X_i)$ 
with  $card(S_{max}) =
card_\uparrow(X)$. Consider the sequence $S_{min}=S_0, S_1,
\ldots,S_n=S_{max}$ where $S_{k+1}$ is the same as $S_k$ except that
$X_{k+1}$ has been assigned its value in $S_{max}$ instead of its
value in $S_{min}$. $|card(S_{k+1})-card(S_k)|\leq 1$
because they only differ on $X_{k+1}$. Hence, 
for any $p\in [card_\downarrow(X),card_\uparrow(X)]$, there
exists $k\in 1..n$ with $card(S_k)=p$.  Thus,  $(S_k,p)$ is a bound
support for $p$ on $\nvalue([X_1, \ldots, X_n], N)$. Therefore, $min(N)$ and
$max(N)$ have a  bound support. 
\myqed

We now prove that
decomposing the \nvalue constraint into
\AtMostNValue and \AtLeastNValue constraints
does not hinder 
pruning when enforcing BC.

\begin{mytheorem}
\label{t:decom_nvalue}
BC on $\nvalue([X_1, \ldots, X_n], N)$ is equivalent to
BC on $\AtMostNValue([X_1, \ldots, X_n], N)$ and  
on $\AtLeastNValue([X_1, \ldots, X_n], N)$. 
\end{mytheorem}
\myproof
  Suppose the \AtMostNValue and \AtLeastNValue constraints
  are BC.  The
  \AtMostNValue constraint guarantees that  $card_\downarrow(X)\leq min(N)$ and the \AtLeastNValue
  constraint guarantees that $card_\uparrow(X) \geq max(N)$.  Therefore, $D(N) \in
  [card_\downarrow(X),card_\uparrow(X)]$. By Lemma \ref{nvalue:prop_1}, the variable $N$ is bound consistent.

Consider a variable/bound value pair $X_i = b$. Let $(S_{least}^b,p_1)$ be a
bound  support of $X_i = b$ in  the \AtLeastNValue constraint and
$(S_{most}^b,p_2)$  be a bound  support of $X_i = b$ in  the \AtMostNValue
constraint. We  have $card(S_{least}^b)\geq p_1$ and
$card(S_{most}^b)\leq p_2$ by definition of \AtLeastNValue and
\AtMostNValue . 
Consider the sequence $S_{least}^b=S^b_0, S^b_1,
\ldots,S^b_n=S_{most}^b$ where $S^b_{k+1}$ is the same as $S^b_k$ except that
$X_{k+1}$ has been assigned its value in $S_{most}^b$ instead of its
value in $S_{least}^b$. $|card(S^b_{k+1})-card(S^b_k)|\leq 1$
because they only differ on $X_{k+1}$. Hence, there 
exists $k\in 1..n$ with $min(p_1,p_2)\leq card(S^b_k)\leq max(p_1,p_2)$.  
We know that $p_1$ and $p_2$ belong to $range(N)$ because they belong
to bound supports. Thus,  $card(S^b_k)\in range(N)$ and
$(S^b_k,card(S^b_k))$ is a bound support for $X_i=b$ on $\nvalue([X_1,
  \ldots, X_n], N)$.  
%
%
  \myqed

When enforcing domain consistency, Bessiere {\it et al.} \cite{bhhkwconstraint2006} noted that decomposing 
the \nvalue constraint into
\AtMostNValue and \AtLeastNValue constraints
does hinder propagation, but only when $dom(N)$ contains 
just $card_\downarrow(X)$ and 
$card_\uparrow(X)$ and there is a gap in the domain
in-between (see Theorem 1 in \cite{bhhkwconstraint2006}
and the discussion that follows). When enforcing BC, 
any such gap in the domain for $N$ is ignored.

\section{\AtMostNValue constraint}\label{sec:atmost}

We now give a decomposition for the \AtMostNValue 
constraint which does not hinder bound consistency propagation.
To decompose the \AtMostNValue constraint, we 
introduce 0/1 variables, $A_{ilu}$ to represent
whether $X_i$ uses a value in the interval
$[l,u]$, and ``pyramid'' variables, $M_{lu}$
with domains $[0, \min\left(u-l+1, n\right)]$ 
which count the number of values
taken inside the interval $[l,u]$. 
To constrain these introduced variables,
we post the following constraints:
\begin{eqnarray}
&  A_{ilu}  = 1 \iff X_i \in [l, u] & \ \ \ \forall \; 1 \leq i \leq n, 1 \leq l \leq u \leq d \label{eqn::firstAtMostNValue} \\
&  A_{ilu}  \leq M_{lu} & \ \ \ \forall \; 1 \leq i \leq n, 1 \leq l \leq u \leq d \label{eqn::lb_AtMostNValue} \\
  & M_{1u}  = M_{1k} + M_{(k+1)u} & \ \ \ \forall \; 1 \leq k < u \leq d \label{eqn::pyram_AtMostNValue}\\
 & M_{1 d}  \leq N & \label{eqn::lastAtMostNValue}
\end{eqnarray}

\begin{myexample}
Consider the decomposition of
an \AtMostNValue constraint over
the following variables and values:
$$
{\scriptsize
\begin{array}{c|ccccc} 
 & 1 & 2 & 3 & 4 & 5  \\ \hline
X_1 & \ast & \ast & \ast & & \ast  \\ 
X_2 & & \ast & & & \\ 
X_3 & & \ast & \ast & \ast & \\ 
X_4 & & & & \ast & \\
X_5 & & & \ast & \ast & \\
N & \ast & \ast  & & & 
\end{array}
}
$$
Observe that we consider that value  5 for $N$ has already been pruned
by \AtLeastNValue, as will be shown in next sections. 
Bound consistency reasoning on the
decomposition will make the following
inferences. As $X_2=2$, 
from \eqref{eqn::firstAtMostNValue} we get $A_{222}=1$.
Hence by \eqref{eqn::lb_AtMostNValue}, $M_{22}=1$.
Similarly, as $X_4=4$, we get $A_{444}=1$ and $M_{44}=1$. 
Now $N\in \{1,2\}$. By \eqref{eqn::lastAtMostNValue}
and \eqref{eqn::pyram_AtMostNValue},
$M_{15} \leq N$,
$M_{15}=M_{14}+M_{55}$, $M_{14}=M_{13}+M_{44}$,
$M_{13}=M_{12}+M_{33}$, $M_{12}=M_{11}+M_{22}$.
Since $M_{22}=M_{44}=1$, we deduce that $N > 1$
and hence $N=2$. This gives
$M_{11}=M_{33}=M_{55}=0$. 
By \eqref{eqn::lb_AtMostNValue}, 
$A_{111}=A_{133}=A_{155}=A_{533}=0$.
Finally, from \eqref{eqn::firstAtMostNValue},
we get $X_1=2$ and $X_5=3$. This gives us bound consistent
domains for the \AtMostNValue constraint.
\end{myexample}

We now prove that this decomposition 
does not hinder propagation in general. 

\begin{mytheorem}
  BC on constraints (\ref{eqn::firstAtMostNValue})
  to (\ref{eqn::lastAtMostNValue}) is equivalent to BC on 
  \AtMostNValue$([X_1,\ldots,X_n], N)$, and takes $O(nd^3)$ time to enforce
down the branch of the search tree. 
\end{mytheorem}
\myproof
\sloppy
First note that changing the domains of the $X$ variables cannot
affect the upper bound of $N$ by the \AtMostNValue constraint and,
conversely, changing the lower bound of $N$ cannot affect the
domains of the $X$ variables. 
  
Let $Y = \{X_{p_1}, \ldots, X_{p_k}\}$ be a maximum cardinality 
subset of variables of $X$ whose ranges are pairwise disjoint (i.e.,
$range(X_{p_i})\cap range(X_{p_j})=\emptyset,\forall i,j\in 1..k,i\neq
j$). Let 
$I_Y = \{[b_i,c_i]\mid b_i = min(X_{p_i}), \ c_i = max(X_{p_i}),
X_{p_i} \in Y\}$ be the corresponding ordered set of disjoint ranges
of the variables in $Y$.  
It has been shown in~\cite{bhhkwcpaior2005} that $|Y|= card_{\downarrow}(X)$. 

Consider the interval $[b_i,c_i] \in I_Y$. Constraints (\ref{eqn::lb_AtMostNValue}) ensure that the variables 
$M_{b_i c_i}$ $i=[1,\ldots,k]$ are greater than or equal to $1$ 
and constraints (\ref{eqn::pyram_AtMostNValue}) ensure that the variable
$M_{1 d}$ is greater than or equal to the sum of lower bounds
of variables $M_{b_i c_i}$, $i=[1,\ldots,k]$, because intervals
$[b_i,c_i]$ are disjoint. 
Therefore, the variable $N$ is greater than or equal to   $card_\downarrow(X)$ and it is bound consistent.

We show that when $N$ is BC and $dom(N)\neq \{card_\downarrow(X)\}$, all
$X$ variables are $BC$. 
Take any assignment $S\in \Pi_{X_i\in X} range(X_i)$ 
such that $card(S)=card_\downarrow(X)$. Let
$S[X_i\gets b]$ be the assignment $S$ where the value of $X_i$ in $S$
has been replaced by $b$, one of the bounds of $X_i$. 
We know that $card(S[X_i\gets b])\in [card(S)-1, card(S)+1]
=[card_\downarrow(X)-1, card_\downarrow(X)+1]$ because
only one variable has been flipped. Hence, any assignment 
$(S,p)$ with $p\geq card_\downarrow(X)+1$ is a bound support. $dom(N)$
necessarily contains such a value $p$ by assumption. 

The only case
when pruning might occur is if the variable $N$ is ground and
$card_\downarrow(X) = N$.  
Constraints (\ref{eqn::pyram_AtMostNValue}) imply that $M_{1d}$ equals the sum of variables
$M_{1,b_1-1} + M_{b_1,c_1} + M_{c_1+1,b_2-1} \ldots + M_{b_N,c_N} + M_{c_N+1,d}$. 
The lower bound of the variable $M_{c_i,b_i}$ is greater than one and
there are $|Y| =card_\downarrow(X)= N$ of these  
intervals. Therefore, by constraint \eqref{eqn::lastAtMostNValue}, the upper bound of variables $M_{c_{i-1}+1,b_i -1}$
that correspond to intervals outside the set $I_Y$ are forced to zero.

%
%
%

  There are $O(nd^2)$ constraints (\ref{eqn::firstAtMostNValue}) and
  constraints (\ref{eqn::lb_AtMostNValue}) that can be
  woken $O(d)$ times down the branch of the search tree. Each requires
    $O(1)$ time for 
  a  total of $O(nd^3)$
  down the branch. 
  There are $O(d^2)$ constraints
  (\ref{eqn::pyram_AtMostNValue}) which can be woken $O(n)$ times  down the
  branch and each invocation takes $O(1)$ time. 
  This gives  a  total of $O(nd^2)$.  
  The final complexity down
  the branch of the search tree is therefore $O(nd^3)$. 
\myqed

\section{Faster decompositions}\label{sec:atmost:faster}

We can improve how our solver handles this decomposition
of the \AtMostNValue constraint
by adding implied constraints and by 
implementing specialized propagators.
Our first improvement is to add an implied constraint
and enforce BC on it:
\begin{eqnarray}
M_{1d} & = & \sum_{i=1}^d M_{ii}
\end{eqnarray}
This does not change the asymptotic
complexity of reasoning with the decomposition,
nor does it improve the level 
of propagation achieved.
However, we have found that the fixed point of 
propagation is reached quicker in practice
with such an implied constraint.

Our second improvement decreases the 
asymptotic complexity of enforcing 
BC on the decomposition of Section \ref{sec:atmost}. 
The complexity  is dominated by reasoning with constraints
\eqref{eqn::firstAtMostNValue} which channel
from $X_i$ to $A_{ilu}$ and thence onto $M_{lu}$ (through constraints
\eqref{eqn::lb_AtMostNValue}). 
If  constraints \eqref{eqn::firstAtMostNValue} were not woken
uselessly,  enforcing BC  should cost $O(1)$  per constraint down the
branch. 
Unfortunately, existing
solvers wake up such constraints as soon as a bound is modified, 
thus a cost in $O(d)$. 
We therefore implemented
a specialized propagator to channel
between $X_i$ and $M_{lu}$ efficiently. 
To be more precise,
we remove the $O(nd^2)$ variables $A_{ilu}$ and replace them 
with $O(nd)$ Boolean variables $Z_{ij}$.
We then add the following constraints

\begin{align}
  Z_{ij} = 1 \iff &X_i \leq j &
  1 \leq j \leq d
  \label{eq:channel-bounds-Z}\\
  Z_{i(l-1)}=1 \vee Z_{iu} = 0 \; \vee  & \; M_{lu} > 0 &
  1 \leq l \leq u \leq d, 1 \leq i \leq n 
  \label{eq:channel-bounds}
\end{align}

These constraints are enough to channel changes in the bounds of the
$X$ variables to $M_{lu}$. There are $O(nd)$ constraints
\eqref{eq:channel-bounds-Z}, each of which can be propagated
in time $O(d)$ over a branch, for a total of $O(nd^2)$.
There are
$O(nd^2)$ clausal
constraints \eqref{eq:channel-bounds}
and each of them can be made 
BC in time $O(1)$ down a
branch of the search tree, for a total cost of $O(nd^2)$. Since 
channeling
dominates
the asymptotic complexity of the entire decomposition 
of Section \ref{sec:atmost}, this improves
the complexity of this decomposition to $O(nd^2)$.
This is similar to the technique used in \cite{bknqwijcai09} to improve
the asymptotic complexity of the decomposition of the \alldiff constraint.

Our third improvement is to enforce stronger pruning by observing that
when $M_{lu}=0$, we can remove the interval $[l,u]$ from all variables,
regardless of whether this modifies their bounds. This corresponds to  
enforcing RC on constraints \eqref{eqn::firstAtMostNValue}. 
Interestingly, this is sufficient to achieve RC on the \AtMostNValue
constraint. 
Unfortunately,  constraints \eqref{eq:channel-bounds} cannot achieve
this pruning and using constraints \eqref{eqn::firstAtMostNValue} 
increases the complexity of the decomposition back to $O(nd^3)$.
We do it by extending the
decomposition
with $O(d\log d)$ Boolean
variables $B_{il(l+2^k)} \in [0,1], 1 \leq i \leq n, 
1 \leq l \leq d, 0 \leq k \leq \lfloor \log d \rfloor$. 
The 
following constraint ensures that $B_{ijj} = 1 \iff X_i = j$. 

\begin{align}
  {\textrm{\sc DomainBitmap}}(X_i, [B_{i11}, \ldots, B_{idd}])
  \label{eq:DomainBitmap}
\end{align}

Clearly we can enforce RC on this constraint in time $O(d)$ over a
branch, and $O(nd)$ for all variables $X_i$. 
We can then use the following
clausal constraints to channel from variables $M_{lu}$
to these variables and on to the $X$ variables.
These constraints are posted for every $1 \leq i \leq n, 1 \leq l \leq u 
\leq d,
1 \leq j \leq d$ and integers $k$
such that $0 \leq k \leq \lfloor \log d \rfloor$:  
\begin{align}
  B_{ij(j+2^{k+1}-1)} = 1 &\vee B_{ij(j+2^{k}-1)}=0 \label{eq:channel-B-first}\\
  B_{ij(j+2^{k+1}-1)} = 1 & \vee B_{i(j+2^k)(j+2^{k+1}-1)}=0 
  \label{eq:channel-B-last} \\
  M_{lu} \neq 0 &\vee B_{il(l+2^{k}-1)} = 0 
  & 2^k \leq u-l+1 < 2^{k+1}
  \label{eq:channel-range-first}\\
  M_{lu} \neq 0 &\vee B_{i(u-2^k+1)u} = 0
  & 2^k \leq u-l+1 < 2^{k+1}
  \label{eq:channel-range-last}
\end{align}

The variable $B_{il(l+2^k-1)}$, similarly to the variables $A_{lu}$, is
true when  $X_i \in [l, l+2^k-1]$, but instead of having one
such variable for every interval, we only have them for 
intervals whose length is a power of two.
When $M_{lu} = 0$, with $2^k \leq u-l+1 < 2^{k+1}$, 
the constraints
(\ref{eq:channel-range-first})--(\ref{eq:channel-range-last})
set to 0 the 
$B$ variables that correspond to the two intervals of length $2^k$
that start at $l$ and finish at $u$, respectively. In turn,
the constraints
(\ref{eq:channel-B-first})--(\ref{eq:channel-B-last})
set to 0 the $B$ variables that correspond to intervals of 
length $2^{k-1}$, all the way down to intervals of size 1. These trigger
the constraints \eqref{eq:DomainBitmap}, 
so all values in the interval $[l,u]$ are
removed from the domains of all variables.

\begin{myexample2}
  Suppose $X_1 \in [5, 9]$. Then, by \eqref{eq:channel-bounds-Z}, 
  $Z_{14} = 0$, $Z_{19} = 1$ and by \eqref{eq:channel-bounds}, $M_{59}>0$. 
  Conversely, suppose $M_{59} = 0$ and $X_1 \in [1, 10]$. Then, by 
  (\ref{eq:channel-range-first})--(\ref{eq:channel-range-last}),
  we get $B_{158} = 0$ and $B_{169} = 0$. 
  From $B_{158} = 0$ and
  (\ref{eq:channel-B-first})--(\ref{eq:channel-B-last})
  we get $B_{156} = 0$, $B_{178} = 0$, $B_{155} = B_{166} = 
  B_{177} = B_{188} = 0$, and by \eqref{eq:DomainBitmap}, the interval
  $[5,8]$ is pruned from $X_1$. Similarly, $B_{169}=0$ causes
  the interval $[6,9]$ to be removed from $X_1$, so
  $X_1 \in [1,4] \cup \{ 10 \}$.
\end{myexample2}

Note that RC can be enforced on each of these constraints in 
constant time over a branch. There exist $O(nd\log d)$ of the constraints
\eqref{eq:channel-B-first}--\eqref{eq:channel-B-last} 
and $O(nd^2)$ of the
constraints~\eqref{eq:channel-range-first}--\eqref{eq:channel-range-last}, 
so the total time to propagate them all
down a branch is $O(nd^2)$.

\section{\AtLeastNValue constraint}

There is 
a similar decomposition for the
\AtLeastNValue constraint. 
We introduce 0/1 variables, $A_{ilu}$ to represent
whether $X_i$ uses a value in the interval
$[l,u]$, and 
integer variables, $E_{lu}$ with
domains $[0,n]$ to count 
the number of times values in $[l,u]$ are \emph{re}-used, that is, how
much the number of variables taking values in $[l,u]$ exceeds the
number $u-l+1$ of values in $[l,u]$. 
To constrain
these introduced variables, we post the following constraints:
\begin{eqnarray}
  & A_{ilu} = 1 \iff X_i \in [l, u] & \ \ \ \forall \; 1 \leq i \leq n, 1 \leq l \leq u \leq d \label{eqn::atleastnvalue-1} \\
 & E_{lu}  \geq \sum_{i=1}^n A_{ilu} - (u-l+1) & \ \ \ \forall \; 1\leq l \leq u \leq d 
 \label{eqn::atleastnvalue-e}\\
&  E_{1u} = E_{1k} + E_{(k+1)u} & \ \ \ \forall \; 1 \leq k < u \leq d \label{eqn::pyram_AtLeastNValue} \\
&  N\leq n-E_{1 d} \label{eqn::atleastnvalue-2}
\end{eqnarray}
\begin{myexample}
Consider the decomposition of an \AtLeastNValue constraint over
the following variables and values:
$$
{\scriptsize
\begin{array}{c|ccccc} 
 & 1 & 2 & 3 & 4 & 5  \\ \hline
X_1 & \ast & \ast & \ast & & \ast  \\ 
X_2 & & \ast & & & \\ 
X_3 & & \ast & \ast & \ast & \\ 
X_4 & & & & \ast & \\
X_5 & & & \ast & \ast & \\
N & \ast & \ast  & & & \ast 
\end{array}
}
$$
Bound consistency reasoning on the
decomposition will make the following
inferences. As $dom(X_i)\subseteq [2,4]$ for $i\in 2..5$, 
from \eqref{eqn::atleastnvalue-1} we get $A_{i24}=1$ for $i\in 2..5$.
Hence, by \eqref{eqn::atleastnvalue-e}, $E_{24}\geq 1$. 
By \eqref{eqn::pyram_AtLeastNValue}, 
$E_{15}=E_{14}+E_{55}$, $E_{14}=E_{11}+E_{24}$. Since $E_{24}\geq 1$
we deduce that  $E_{15}\geq 1$. 
Finally, from \eqref{eqn::atleastnvalue-2} and the fact that  $n=5$,
we  get $N\leq 4$.  This gives us bound consistent
domains for the \AtLeastNValue constraint.
\end{myexample}

We now prove that this decomposition
does not hinder propagation in general. 

\begin{mytheorem}
  \label{thm:atleastnvalue-bc}
  BC on the constraints
  \eqref{eqn::atleastnvalue-1} 
  to \eqref{eqn::atleastnvalue-2} is equivalent to BC
  on \AtLeastNValue$([X_1,\ldots,X_n], N)$, and takes $O(nd^3)$ time to enforce
down the branch of the search tree.
\end{mytheorem}

\myproof
  First note that changing the domains of the $X$ variables cannot
  affect the lower bound of $N$ by the \AtLeastNValue constraint and,
  conversely, changing the upper bound of $N$ cannot affect the
  domains of the $X$ variables. 
  
  It is known \cite{bcp01} that $card_{\uparrow}(X)$ is equal to the
  size of a maximum matching $M$ in the value graph of the
  constraint. Since $N\leq n-E_{1d}$, we show that the lower bound of
  $E_{1d}$ is equal to $n-|M|$.\footnote{We assume that $E_{1d}$
    is not pruned by other constraints.}
  We first show that we can construct a matching $M(E)$ of size
  $n-min(E_{1d})$, then show that it is a maximum matching. The proof uses 
  a partition of the interval $[1,d]$ into a set of maximal
  saturated intervals $I = \{[b_j,c_j]\}$, $j=1,\ldots,k$ such that $min(E_{b_j, c_j}) = \sum_{i=1}^n min(A_{ib_jc_j}) -
  (c_j-b_j+1)$ and a set of unsaturated intervals $\{[b_j,c_j]\}$such that 
  $min(E_{b_j, c_j}) = 0$.

  Let $I=\{ [b_j, c_j] \mid j \in [1\ldots k] \}$ be the ordered set of maximal
  intervals such that $min(E_{b_j, c_j}) = \sum_{i=1}^n min(A_{ib_jc_j}) -
  (c_j-b_j+1)$. Note that the intervals in $I$ are disjoint otherwise 
  intervals are not maximal. An interval $[b_i,c_i]$ is smaller than
  $[b_j,c_j]$ iff $c_i < b_j$. We denote the union of the first $j$ intervals  $D_I^j = \bigcup_{i=1}^j [b_i,c_i]$,
  $j =[1,\ldots,k]$, $p = |D_I^k|$ and
  the variables whose domain is inside one of intervals $I$ 
  $X_I=\{X_{p_i}| D(X_{p_i}) \subseteq D_I^k\}$. 
  
  Our construction of a matching uses two sets of
  variables, $X_I$ and $X \setminus X_I$. First, we identify the cardinality of these
  two sets. Namely, we show that the size of the set $X_I$ is  $p + min(E_{1,d})$
  and the size of the set $X\setminus X_I$ is $n - (p + min(E_{1,d}))$.
  
  Intervals $I$ are saturated therefore each value from these intervals are
  taken by a variable in $X_I$. Therefore, $X_I$ has size at least $p$.
  Moreover, there exist $min(E_{1d})$ additional variables that take values from $D_I^k$,
  because values from intervals between two consecutive intervals in $I$ 
  do not contribute to the lower bound of the variable $E$ by construction of $I$.
  Therefore, the number of variables in $D_I^k$ is at least $p + min(E_{1,d})$.
  Note 
  that constraints \eqref{eqn::pyram_AtLeastNValue} imply that $E_{1d}$ equals 
  the sum of variables $E_{1,b_1-1} + E_{b_1,c_1} + E_{c_1+1,b_2-1} \ldots + E_{b_k,c_k} + E_{c_k+1,d}$.
  As intervals in $I$ are disjoint then $\sum_{i=1}^k min(E_{b_i,c_i}) = |X_I| - p$. 
  If $|X_I| > p + min(E_{1,d})$ then  $\sum_{i=1}^k min(E_{b_i,c_i}) >  min(E_{1,d})$ and the lower
  bound of the variable $E_{1d}$ will be increased. Hence, $|X_I| = p + min(E_{1,d})$.
  
  Since all these intervals are saturated, we can
  construct a matching $M_I$ of size $p$ using the variables in
  $X_I$. The size of $X \setminus X_I$ is $n-p-min(E_{1d})$. We show by
  contradiction that we can construct a matching $M_{D-D^k_I}$ of size
  $n-p-min(E_{1d})$ using the variables in $X \setminus X_I$ and the
  values $D-D_I^k$.

  Suppose such a matching does not exist. Then, there exists an
  interval $[b, c]$ such that $|(D \setminus D_I^k) \cap [b,c]| < \sum_{i
    \in X \setminus X_I} min(A_{ibc})$, i.e., after consuming the values in
  $I$ with variables in $X_I$, we are left with fewer values in
  $[b,c]$ than variables whose domain is contained in $[b,c]$.  We
  denote $p' = |[b,c] \cap D^k_I|$,
  so that $p'$ is the number of values inside the interval $[b,c]$
  that are taken by variables in $X_I$. The total number of variables
  inside the interval $[b,c]$ is greater than or equal to
  $\sum_{i=1}^n min(A_{ibc})$. The total number of variables $X_I$ inside
  the interval $[b,c]$ equals to $p' + min(E_{b,c})$. 
  Therefore, $\sum_{i \in  X \setminus X_I} min(A_{ibc}) \leq \sum_{i=1}^n min(A_{ibc}) - p' -
  min(E_{b,c})$. On the other hand, the number of values that are not
  taken by the variables $X_I$ in the interval $[b,c]$ is $c-b+1 -
  p'$. Therefore, we obtain the inequality $c-b+1 -p' < \sum_{i=1}^n
  min(A_{ibc}) - p' - min(E_{b,c})$ or $ min(E_{bc}) < \sum_{i=1}^n min(A_{ibc}) -
  (c-b+1) $.  By construction of $I$, $\sum_{i=1}^n min(A_{ibc}) - (c-b+1)
  < min(E_{bc})$, otherwise the intervals in $I$ that are subsets of
  $[b,c]$ are not maximal. This leads to a contradiction, so we can
  construct a matching $M(E)$ of size $n-min(E_{1d})$.

  Now suppose that $M(E)$ is not a maximum matching.  This means
  that $min(E_{1d})$ is overestimated by propagation on
\eqref{eqn::atleastnvalue-1} 
  and \eqref{eqn::atleastnvalue-2}.  Since
  $M(E)$ is not a maximum matching, there exists an augmenting
  path of $M(E)$, that produces $M'$, such that $|M'| =
  |M(E)|+1$. This new matching covers all the values that
  $M(E)$ covers and one additional value $q$. We show that $q$
  cannot belong to the interval $[1,d]$. 

	The value $q$ cannot be in any   interval in $I$, 
	because all values in $[b_i,c_i] \in I$ are used by variables whose domain is contained in $[b_i,c_i]$. 
	In addition, $q$ cannot be in an interval $[b,c]$ between two consecutive 
	intervals in $I$, because those intervals do
  not contribute to the lower bound of $E_{1d}$. Thus, $M'$ cannot
  cover more values than $M(E)$ and they must have the same size,
  a contradiction.


We show that when $N$ is BC and $dom(N)\neq \{card_\uparrow(X)\}$, all
$X$ variables are $BC$. 
Take any assignment $S\in \Pi_{X_i\in X} range(X_i)$ 
such that $card(S)=card_\uparrow(X)$. Let
$S[X_i\gets b]$ be the assignment $S$ where the value of $X_i$ in $S$
has been replaced by $b$, one of the bounds of $X_i$. 
We know that $card(S[X_i\gets b])\in [card(S)-1, card(S)+1]
=[card_\uparrow(X)-1, card_\uparrow(X)+1]$ because
only one variable has been flipped. Hence, any assignment 
$(S,p)$ with $p\leq card_\uparrow(X)-1$ is a bound support. $dom(N)$
necessarily contains such a value $p$ by assumption. 

  We now show that if $N=card_\uparrow(X)$, enforcing BC on the constraints
 \eqref{eqn::atleastnvalue-1}--\eqref{eqn::atleastnvalue-2} makes the
  variables $X$ BC with respect to the \AtLeastNValue constraint.
We first observe  that in a  bound support, variables  $X$ must take
the maximum number of different values because $N=card_\uparrow(X)$. 
Hence, in a bound support, variables $X$ 
that 
are not included in a saturated interval
 will take values outside any saturated interval they overlap and they
 all take different values. 
We recall that $min(E_{1d})=n-|M|=n-card_\uparrow(X)$. Hence, by
constraint \eqref{eqn::atleastnvalue-2},  $E_{1d}=n-N$. 
  We recall the the size of set $X_I$ equals $ p + E_{1d}$.  
  Constraints \eqref{eqn::pyram_AtLeastNValue} imply that $E_{1d}$ equals 
  the sum of variables $E_{1,b_1-1} + E_{b_1,c_1} + E_{c_1+1,b_2-1} \ldots + E_{b_k,c_k} + E_{c_k+1,d}$
  and $\sum_{i=1}^k min(E_{b_i,c_i}) =|X_I| - p=
  min(E_{1d})=max(E_{1d}) $. Hence, by
  constraints~\eqref{eqn::pyram_AtLeastNValue}, the upper bounds of
  all variables $E_{b_i,c_i}$ that correspond to the saturated intervals
  are forced to  $min(E_{b_i,c_i})$. 
Thus, by constraints \eqref{eqn::atleastnvalue-1} and
\eqref{eqn::atleastnvalue-e}, all variables in $X\setminus X_I$ have
their bounds pruned if they belong to  $D^k_I$. 
By constraints~\eqref{eqn::pyram_AtLeastNValue} again, 
  the upper bounds of all
  variables $E_{lu}$
  that correspond to the  unsaturated intervals are forced to take
  value 0, and all variables $E_{l'u'}$ with $[l',u']\subseteq[l,u]$
  are forced to 0
  as well. 
Thus, by constraints 
\eqref{eqn::atleastnvalue-1} and 
\eqref{eqn::atleastnvalue-e}, all variables in $X\setminus X_I$ have
their bounds pruned  if they belong to a Hall interval of other
variables in $X\setminus X_I$. This is  what BC on the \alldiff
constraint does \cite {bknqwijcai09}. 

  There are $O(nd^2)$ constraints \eqref{eqn::atleastnvalue-1} that can be
  woken $O(d)$ times down the branch of the search tree in  $O(1)$, so
  a  total of $O(nd^3)$
  down the branch. 
  There are $O(d^2)$ constraints
  \eqref{eqn::atleastnvalue-e} which can be 
  propagated in time $O(n)$
  down the
  branch 
  for a $O(nd^2)$.  
  There are $O(d^2)$ constraints
  \eqref{eqn::pyram_AtLeastNValue} which can be woken $O(n)$ times each down the
  branch 
for a total cost in  $O(n)$ time down the
  branch. Thus  a  total of $O(nd^2)$. 
  The final complexity down
  the branch of the search tree is therefore $O(nd^3)$. 
  \myqed

  The complexity of enforcing BC on the \AtLeastNValue constraint can
  be improved to $O(nd^2)$ in way similar to that described in Section
  \ref{sec:atmost:faster} and in \cite{bknqwijcai09}.

\myOmit{
\section{\usedby constraint}

Another global constraint that can be decomposed in
a similar way is 
$\usedby([X_1, \dots, X_n], [Y_1, \ldots, Y_m])$.
This is satisfied iff the multiset of values
used by $X_i$ is contained within 
the multiset used by $Y_i$ \cite{bktcpaior04}.
Beldiceanu, Katriel and Thiel 
suggest this can be used to model
crew pairing problems, and propose a flow-based
propagator for enforcing DC in 
$O(n^2m)$ time and
BC in $O(m\alpha(m,m)+n\log n)$ time
where $\alpha$ is 
the slow-growing inverse Ackermann function
\cite{bktcpaior04}.  
We are not aware of {\em any} constraint solver which 
supports a propagator for the \usedby constraint.
It is therefore useful to come up with a 
decomposition that enables the \usedby constraint
to be propagated. 

To decompose the \usedby constraint,
we again used a pyramid of constraints
to count values used. In this case, we 
put together two such pyramids: 
\begin{eqnarray}
&  A_{ilu} = 1 \iff X_i \in [l, u] & \ \ \ \forall \; 1 \leq i \leq n, 1 \leq l \leq u \leq d \label{eqn::firstUsedBy}  \\
&  B_{ilu}  = 1 \iff Y_i \in [l, u] & \ \ \ \forall \; 1 \leq i \leq m, 1 \leq l \leq u \leq d \label{eqn::secondUsedBy}\\
&  N_{lu}  = \sum_{i=1}^m B_{ilu} & \ \ \ \forall \; 1 \leq l \leq u \leq d \label{eqn::used-by-blu}\\
&  N_{1u}  = N_{1k} + N_{(k+1)u} & \ \ \ \forall \; 1 \leq k \leq u \leq d \label{eqn::used-by-pyramid}\\
&  \sum_{i=1}^n A_{ilu}  \leq N_{lu} & \ \ \ \forall \; 1 \leq l \leq u \leq d \label{eqn::lastUsedBy}
\end{eqnarray}

\begin{myexample}
Consider the decomposition of a \usedby constraint over
the following variables and values:
$$
{\scriptsize
\begin{array}{c|ccccc} 
 & 1 & 2 & 3 & 4 & 5  \\ \hline
X_1 & \ast & \ast &  & &   \\ 
X_2 & \ast & \ast & & & \\ 
X_3 & &  &  & \ast & \ast\\ 
X_4 & & & & \ast & \ast\\
X_5 & \ast&\ast & \ast & \ast & \ast
\end{array}
\hspace{2em}
\begin{array}{c|ccccc} 
 & 1 & 2 & 3 & 4 & 5  \\ \hline
Y_1 & \ast & \ast & \ast & \ast & \ast  \\ 
Y_2 & \ast & \ast & \ast & \ast & \ast  \\ 
Y_3 & \ast& \ast & \ast & \ast & \ast\\ 
Y_4 & & & \ast& \ast & \ast\\
Y_5 & & & \ast &  & 
\end{array}
}
$$
Bound consistency reasoning on the
decomposition will make the following
inferences. As $dom(X_i)\subseteq [1,2]$ for $i\in 1,2$,  
from \eqref{eqn::firstUsedBy} we get $A_{112}=1$ and $A_{212}=1$.
As $dom(X_i)\subseteq [4,5]$ for $i\in 3,4$,  
from \eqref{eqn::firstUsedBy} we get $A_{345}=1$ and $A_{445}=1$.
Hence, by \eqref{eqn::lastUsedBy}, 
we get $N_{12}\geq 2$ and $N_{45}\geq 2$.
By \eqref{eqn::used-by-pyramid}, 
$N_{15}=N_{13}+N_{45}$, $N_{13}=N_{12}+N_{33}$. Now by 
\eqref{eqn::used-by-blu}, we get $N_{15}\leq 5$. 
Since $N_{12}\geq 2$
and $N_{45}\geq 2$
we deduce that  $N_{33}\leq 1$. 
As $dom(Y_5)=\{3\}$, from \eqref{eqn::secondUsedBy} we get
$B_{533}=1$.  From \eqref{eqn::used-by-blu},
we  get $B_{i33}=0$ for $i\in[1,4]$. Finally, from BC on 
\eqref{eqn::secondUsedBy}, we prune value 3 from $dom(Y_4)$. 
By \eqref{eqn::lastUsedBy} and   $N_{33}\leq 1$ we
This gives us bound consistent
domains for the $Y_j$ on the \usedby constraint.
With similar  propagations, we deduce that $B_{533}=1$, then
$N_{33}\geq 1$, then $N_{12}\leq 2$ and $N_{45}\leq 2$, then
$A_{512}=0$ and $A_{545}=0$, then $X_5=3$. The \usedby constraint
is therefore BC. 
\end{myexample}

We now prove that this decomposition
does not hinder propagation in general.

\begin{mytheorem}
  BC on constraints (\ref{eqn::firstUsedBy})
  to (\ref{eqn::lastUsedBy}) is equivalent to BC on the \usedby\
  constraint, and takes $O(md^3)$ time to enforce down the
branch of the search tree.
\end{mytheorem}
\myproof
  We construct a bipartite graph $G$ with partitions of $n$ and $m$
  vertices, such that there exists a vertex in $A$ for each variable
  $X_i$ and a vertex in $B$ for each variable $Y_j$. We refer to
  vertices by the name of the corresponding variable. There exists an
  edge between $X_i$ and $Y_j$ iff the intervals $[min(X_i), max(X_i)]$
  and $[min(Y_j), max(Y_j)]$ intersect.
  It is easy to see that the \usedby constraint is satisfiable iff
  there exists a matching of size $n$ in $G$. 

  In order to enforce bound consistency, we prune variables in two
  cases. First, if there exists a value $k$ such that $k \notin
  range(Y_j)$ for any $j$, then prune $k$ from  $dom(X_i)$ for all $i$
  such that $k$ is a bound of $X_i$. In
  this case, $B_{jkk} = 0$ for all $j$, therefore constraint
  \eqref{eqn::lastUsedBy} sets $A_{ikk}=0$ for all $i$ and $k$ is
  pruned from the domain of all $X_i$ variables by constraint
  \eqref{eqn::firstUsedBy}  if $k$ is a bound of $X_i$.

  Second, we perform pruning when $G$ contains a Hall
  set~\cite{Hall35}. A Hall set is a set  
  $H = I \cup J$, $|I| =
  |J| = k$ that consists of $X$ variables $\{X_i \mid
  i \in I\}$ and $Y$ variables $\{ Y_j \mid j \in J\}$ such that 
  $\forall p \in I: range(X_p) \subseteq \cup_{q \in J} range(Y_q)$ and
  $\forall p \in I: range(X_p) \cap \left( \cup_{q \notin J} range(Y_q)
  \right) = \emptyset$. 
Then, we need to prune the domains of the $Y$
  variables so that $range(Y_q) \subseteq \cup_{p \in I} range(X_p)$ for all
  $q \in J$ and we need to prune the $X$ variables so that 
$range(X_p) \cap range(Y_q) = \emptyset$ for all $p \notin I,
  q \in J$. We show that the decomposition enforces BC in this case as
  well. Let   $a = \min\cup_{p \in I} D(X_p)$ and $b = \max\cup_{p \in I} D(X_p)$.

  First we split the interval $[a,b]$	into $r$ disjoint subintervals
  $[I_1 S_2 I_3 S_4\ldots S_{r-1}I_r]$
  in such a way that the intervals $I_i$ are subsets of $\cup _{i \in
    I} range(X_i)$, while $S_i$ are the maximum length intervals that
  do not intersect 
$\cup _{i \in I} range(X_i)$. 
 	We denote $s_i$ the  number of variables $Y$ that are contained in the subinterval $S_i$, $i=2,4,\ldots,r-1$. 
 	By  \eqref{eqn::used-by-blu}, $min(N_{S_i}) \geq s_i$.
 	We denote $k_i$ the  number of variables $X$ that are contained in the subinterval $I_i$,  $i=1,3,\ldots,r$. 
 	By  \eqref{eqn::lastUsedBy}, 
$min(N_{I_i}) \geq k_i$.	By the assumption that $[a,b]$ is a Hall set, we conclude that the total number of variables $Y$
 	that overlap the intervals $I_i$, $i=1,3\ldots,r$  is exactly $k$.  Therefore, the total number of
 	variables $Y$ that overlap $[a,b]$ is exactly $k+s$ and $max(N_{ab}) = k + s$.

 	Second we observe that the total number of variables $Y$, $m$, equals the sum of the number of variables that overlap $[a,b]$, which is exactly $k+s$ 
 	and the number of variables $Y$ that do not overlap $[a,b]$, which is exactly the variables
 	that are contained in the intervals $[1,a-1]$ and $[b+1,d]$. Therefore, we
 	get that $m = min(N_{1(a-1)}) + (k+s)+ min(N_{(b+1)d})$.  
 	
 	Consider the partitioning of the interval $[1,d]$ into disjoint intervals $\{[1,a-1], I_1, S_2,\ldots, S_{r-1},I_r, [b+1,d]\}$. 
 	The sum of minimum values of the variables $N_{1(a-1)},
        N_{I_1}, N_{S_2},\ldots, N_{S_{r-1}},N_{I_r}, N_{(b+1)d}$ is
        at least 
 	$m$ because $\sum_i k_i=k$, $\sum_i s_i=s$, $min(N_{I_i})\geq
        k_i$, and $min(N_{S_i})\geq s_i$. By the constraints \eqref{eqn::used-by-pyramid}, all these variables will be set to their minimum value. Consider, for example,
 	the constraint $N_{1d} = N_{1b} + N_{(b+1)d}$.  By  \eqref{eqn::used-by-pyramid}, 
 	we obtain that $min(N_{1b})\geq (min(N_{1(a-1)}) + k + s)$. As $m = min(N_{1(a-1)}) + (k+s)+ min(N_{(b+1)d})$, then 
 	we get that the variables $N_{1b}$ and $N_{(b+1)d}$ are ground.  Now, that variable $N_{1b}$ is ground and we can apply
 	the same argument to fix variables $N_{I_r}, N_{S_{r-1}}, \ldots, N_{S_2}, N_{I_1}, N_{1(a-1)}$.
 	
 	Note that the intervals $[1,a-1], S_2,\ldots, S_{r-1}, [b+1,d]$ are saturated by variables $Y_j$, $j \notin J$,
 	because the corresponding pyramid variables $N$ were fixed to their minimum values. Therefore, these values will be
 	pruned from domains of variables $Y_j$, $j \in J$. Finally, the variables $N_{I_i}$ are fixed to $k_i$, $i=1,3,\ldots,r$.
 	Hence, by constraints \eqref{eqn::lastUsedBy}, values inside the intervals $I_i$ will be pruned from domains of variables $X_i$, $i \notin I$.


  As for constraint (\ref{eqn::firstAtMostNValue}), constraints
  (\ref{eqn::firstUsedBy}) and (\ref{eqn::secondUsedBy}) takes
  $O(nd^3)$ and $O(md^3)$ to propagate down the branch of the search
  tree. There are $O(d^2)$ constraints (\ref{eqn::used-by-blu}). These
  constraints can be propagated in $O(m)$ time down the branch of the
  search tree for a total complexity of $O(md^2)$ down the
  branch. There are $O(d^2)$ constraints (\ref{eqn::used-by-pyramid})
  that can be woken $O(m)$ times. Each propagation requires $O(1)$
  time for a total of $O(md^2)$ down the branch of the search
  tree. There are $O(d^2)$ constraints (\ref{eqn::lastUsedBy}) that
  can each be propagated in $O(n)$ time down the branch of the search
  tree for a total running time of $O(nd^2)$ for this constraint). The
  total running time complexity is therefore dominated by $O(md^3)$.
  \myqed
}

\section{Experimental results}

To evaluate these decompositions,
we performed experiments on two problem domains. 
We used the same problems as in a previous experimental comparison 
of propagators for the $\AtMostNValue$ constraint
\cite{bhhkwconstraint2006}. 
We ran experiments with
Ilog Solver 6.2 on an Intel Xeon 4 CPU, 2.0 Ghz, 4Gb RAM.  

\subsection {Dominating set of the Queen's graph}

The problem is to put the minimum number of queens on a $n \times n$ chessboard,
so that each square either contains a
queen or  is attacked by one. 
This is equivalent to the dominating set problem of the
Queen's graph. 
Each vertex in the Queen's graph corresponds to a square of the chessboard and there exists
an edge between two vertices iff
a queen from one square can attack a queen from the other square. 
To model the problem, we use a variable $X_i$ for each
square, and values from $1$ to $n^2$ and
post a single $\AtMostNValue([X_1,\ldots, X_{n^2}],N)$ constraint. 
The value $j$ belongs to $D(X_i)$ iff
there exists an edge  $(i, j)$ in the Queen's graph or $j=i$. 
We use 
minimum  domain variable ordering and 
a lexicographical value ordering. 
For $n\leq 120$,
all minimum dominating sets for the Queen's problem 
are either of size $\left\lceil n/2\right\rceil$ or $\left\lceil n/2 + 1\right\rceil$
\cite{ostergard}. We therefore only solved instances for
these two values of $N$. 

We compare our decomposition with two simple decompositions of the $\AtMostNValue$ constraint.
The first decomposition 
is the one described in Section \ref{sec:nvalue:simple} except that in
constraint \eqref{dec3}, we replace ``$=$'' by ``$\leq$''. 
We denote this decomposition $Occs$. The second decomposition is similar to the first one, but we use the cardinality
variables of a $\gcc$ constraint to keep track of the used values. We call this decomposition  $Occs_{gcc}$.
The final two decompositions are variants of the decomposition described
in Section \ref{sec:atmost}, which we call
$Pyramid_{BC}$ or $Pyramid_{RC}$ depending whether we
enforce BC  or RC on 
our decomposition. 
As explained in Section \ref{sec:atmost:faster},
we channel the variables $X_i$ directly to the pyramid variables $M_{lu}$ to avoid introducing many
auxiliary variables $A_{ilu}$ and we  add the redundant constraint $\sum_{i=1}^{n^2}M_{ii} = M_{1,n^2}$ to the decomposition 
to speed up the propagation across the pyramid. 
For the decomposition that enforces RC, we did not
fully implement the $O(nd^2)$ decomposition
of Section \ref{sec:atmost:faster}, but rather
a simple channeling propagator that achieves RC in $O(nd^3)$
on \eqref{eqn::firstAtMostNValue},
but with better asymptotic constants than constraints 
\eqref{eqn::firstAtMostNValue}
.
Finally, we  re-implemented the ternary sum constraint $Z = X+Y$ in Ilog. This gave
us about $30\%$ speed up.

\begin{table}[htb]
\begin{center}
{\small
\caption{\label{t:t1} Backtracks and rumtime (in seconds) to
solve the dominating set problem for the Queen's graph. 
Best results for any statistic are bold fonted. 
}
\begin{tabular}{|  c|c ||rr|rr|rr|rr|}
\hline
$n$ & $N$ 
&\multicolumn {2}{|c|}{$Occs$}
&\multicolumn {2}{|c|}{$Occs_{gcc}$}
&\multicolumn {2}{|c|}{$Pyramid_{BC}$}  
&\multicolumn {2}{|c|}{$Pyramid_{RC}$}  \\
\hline 
\multicolumn {2}{|c||}{}
&backtracks & time &backtracks & time &backtracks & time &backtracks & time\\
\hline   

 5 & 3 &       34 &  0.01  
&       34 &  0.06  
 &        \textbf{ 7} &  \textbf{ 0.00}  
&         \textbf{ 7} &  \textbf{ 0.00}  
\\ 
 6 & 3 &      540 &  0.16  
&      540 &  2.56  
&       \textbf{ 118} &  \textbf{ 0.03}  
&       \textbf{ 118} &  \textbf{ 0.03}  
\\ 
 7 & 4&   195,212 & 84.50  
&    195,212 & 1681,21  
&     \textbf{ 83,731} &  \textbf{15.49}  
&    \textbf{ 83,731} & 21.21  
\\ 
 8 & 5 &   390,717 & 255.64  
&   390,717 & 8,568.35  
&    \textbf{ 256,582} &  \textbf{58.42}  
&   \textbf{ 256,582} & 89.30  
\\ 
\hline 

\end{tabular}}
\end{center}
\end{table}

Results are presented in Table~\ref{t:t1}. 
Our decomposition performs better than the other two decompositions,
both in runtime and in  number of backtracks. We observe that BC and
RC  prune the same (i.e., same number of backtracks) on our decomposition
but  BC is faster on larger problems. 
It should be pointed out that our results are comparable with the results for the $\AtMostNValue$ bounds consistency
propagator from~\cite{bhhkwconstraint2006}.
Whilst our decomposition is not as efficient as the best results 
presented in that paper, our decomposition was easier to implement.

\subsection {Random binary CSP problems}
We also reproduced the set of experiments on random binary CSP problems from ~\cite{bhhkwconstraint2006}.
These problems can be described by four parameters. The number of variables $n$, the domain size $d$,
the number of binary constraints $m$ and the number of forbidden tuples in each binary constraint.  
The first three classes are hard problems at the 
phase transition in satisfiability. 
The last two classes are under-constrained problems.
We add a single $\AtMostNValue$ constraint over all variables to bound the number of values $N$ that can be used in a solution.

As in \cite{bhhkwconstraint2006}, 
we generated 500 instances for each of the following 5 classes:
\begin{itemize}
	\item class A : $n = 100, d = 10, m = 250, t = 52, N = 8$
	\item  class B : $n = 50, d = 15, m = 120, t = 116, N = 6$
	\item  class C : $n = 40, d = 20, m = 80, t = 240, N = 6$
	\item class D : $n = 200, d = 15, m = 600, t = 85, N = 8$
	\item class E : $n = 60, d = 30, m = 150, t = 350, N = 6$
\end{itemize}

All instances are solved using the minimum domain variable ordering heuristic, a lexicographical value 
ordering and a timeout of $600$ seconds.  We use the same decompositions of
the \AtMostNValue constraint as in the experiments with the dominating 
set of the Queen's graph. Results are given in Table~\ref{t:t2}.  
On classes $A,B,C$ (hard problems), our decomposition  is faster than
the other two decompositions and solves more instances whatever we use
BC or RC
. 
On classes $D,E$ (under-constrained problems),  enforcing BC 
on our decomposition
does not prune 
the search space enough. This leads to a high number of backtracks and a
significant slow down. Our decomposition with RC 
is again better than the other decompositions.

\begin{table}[tb]
\begin{center}
{\small
\caption{\label{t:t2} Randomly generated binary CSPs with an $\AtMostNValue$ constraint. 
For each class we give two lines of results. Line 1:
number of instances solved in 600 sec (\#solved),  average backtracks on solved
instances (\#bt),  average time on solved instances (time). Line 2: 
number of instances solved by all methods,  
average backtracks and time on these instances. 
Best results for any statistic are bold fonted. 
}
\begin{tabular}{|  cr |rrr|rrr|rrr|rrr|}
\hline
$$ 
&&\multicolumn {3}{|c|}{$Occs$}
&\multicolumn {3}{|c|}{$Occs_{gcc}$}
&\multicolumn {3}{|c|}{$Pyramid_{BC}$}
&\multicolumn {3}{|c|}{$Pyramid_{RC}$}
 \\
\hline 
\hline
$Class$ 
&&\multicolumn {3}{|c|}{\#solved ~~ \#bt ~~ time}
&\multicolumn {3}{|c|}{\#solved ~~ \#bt ~~ time}
&\multicolumn {3}{|c|}{\#solved ~~ \#bt ~~ time}
&\multicolumn {3}{|c|}{\#solved ~~ \#bt ~~  time}\\
\hline 
 A  & total solved & 453 & 139,120 &  111.2 & 79 &   8,960 &  302.8 & \textbf{463}&  \textbf{ 168,929} & \textbf{  101.8} & {462}&  { 148,673} & {  105.7} \\ 
&  solved by all & 79 & 8,960 & 7.1 & 79 & 8,960 &  302.8& 79 &   \textbf{   9,104} & \textbf{    5.7}&  79 &  {   8,739} & {    6.3} \\ 
\hline 
B  & total solved & 473 & 228,757 & 113.5 & 125 &  37,377 &  292.9 & \textbf{492}&  \textbf{ 224,862} & \textbf{ 89.0} & {491}& { 235,715} & {   94.9} \\ 
& solved by all & 125 & 7,377 & 17.6 & 125 & 37,377 &  292.9 &125 & \textbf{  32,810} & \textbf{   10.9} &125 & {  32,110} & {   12.2} \\ 
\hline 
 C  & total solved & 479 & 233,341 & 110.3 & 156 &  37,242 & 290.3 & \textbf{492}&  \textbf{ 234,915} & \textbf{   79.5} & {490}& { 224,802} & {   84.2} \\ 
& solved by all & 156 & 37242 &  16.4 & 156 &   37,242 & 290.3 &  156 & \textbf{  32,184} & \textbf{   9.7} &  156 & {  31,715} & {   11.1} \\ 
\hline 
\hline 
 D & total solved & 482 &   8,306 & 6.0 & 456 &    207 &   14.9 & {416}&  {  168,021} & {    24.2} & \textbf{489}&  \textbf{  13,776} & \textbf{    9.0} \\ 
 & solved by all & 391 & \textbf{    195} & \textbf{    0.2} & 391 & 195 & 13.1 &  391 & 145,534 & 14.9 & 391 & 690 & 0.4 \\ 
\hline 
 E & total solved & \textbf{500}&    331 & 0.3 & \textbf{500}&   331 & 5.1 & \textbf{500}&  {    4,252} & {    0.4} & \textbf{500}&  \textbf{    174} & \textbf{    0.1} \\ 
 & solved by all  & 500 & 331 & 0.3 & 500 & 331 & 5.1 & 500 & {    4,252} & {    0.4} & 500 & \textbf{    174} & \textbf{    0.1} \\
\hline 
 \multicolumn {2}{|c|} { TOTALS }& & & & & & & & & &&&\\ 
 \multicolumn {2}{|r|} {Total solved/tried}& \multicolumn {3}{|c|} { 2,387/2,500}& \multicolumn {3}{|c|} { 1,316/2,500}& \multicolumn {3}{|c|} {{2,363}  /2,500}& \multicolumn {3}{|c|} {\textbf{2,432}/2,500}\\ 
 \multicolumn {2}{|r|} {Avg time for solved}& \multicolumn {3}{|c|}{ 67.0} & \multicolumn {3}{|c|}{ 87.5}& \multicolumn {3}{|c|} {59.364} & \multicolumn {3}{|c|}{\textbf{ 58.0}} \\ 
 \multicolumn {2}{|r|} {Avg bts for solved}& \multicolumn {3}{|c|}{ 120,303} & \multicolumn {3}{|c|}{   8,700} & \multicolumn {3}{|c|} {163,473} &\multicolumn {3}{|c|}{\textbf{ 123,931}} \\ 
\hline 
\end{tabular}}
\end{center}
\end{table}

These experiments demonstrate that 
this new decomposition is efficient to use
in practice. Of course, if the toolkit contains a specialized 
BC propagator for the \nvalue constraint, we will probably do best
to use this. However, when the toolkit lacks such
a propagator (as is often the case), it is reasonable
to try out our decomposition.

\section{Other related work}

Decompositions have been given for
a number of other global constraints. 
For example, Beldiceanu {\it et al.} identify conditions
under which global constraints specified
as automata can be decomposed into signature
and transition constraints without hindering
propagation \cite{bcdpconstraints05}. 
As a second example, many global constraints
can be decomposed using \roots and \range
which can themselves often be 
propagated effectively 
using simple decompositions 
\cite{bhhkwijcai2005,bhhkwcp2006,bhhkwcpaior06}. 
As a third example, 
the \regular and \grammar constraints can be decomposed
without hindering propagation
\cite{qwcp06,qwcp07,bhhkwecai08}. 
As a fourth example,
decompositions of the \sequence constraint 
have been shown to be effective
\cite{bnqswcp07}. 
Most recently, we demonstrated that
the \alldiff and \gcc constraint
can be decomposed into simple primitive constraints
without hindering bound consistency propagation 
\cite{bknqwijcai09}. These decompositions
also introduced variables to count variables
using values in an interval. For example, the
decomposition of \alldiff ensures that
no interval has more variables taking 
values in the interval than the number of values in
the interval. 
Using a circuit complexity
lower bound, we also proved that
there is no polynomial sized SAT decomposition
of the \alldiff constraint (and therefore 
of its generalizations like \nvalue) on 
which unit propagation achieves
domain consistency \cite{bknwijcai09}.

\section{Conclusions}

We have shown that the \nvalue
constraint can be decomposed 
into simple arithmetic  constraints. This decomposition
permits a global view to be maintained
that achieves bound consistency. Our experiments
demonstrate that this decomposition is
a relatively efficient and effective means to propagate 
the global \nvalue constraint. 
Decompositions of global constraints like this
are interesting for a number
of other reasons. First
they provide fresh insight into the
workings of specialized propagation algorithms.
In this case, it is surprising that interval graph
reasoning used to enforce bound consistency on the 
\nvalue can be simulated with simple arithmetic constraints. 
Second, such decompositions may make nogood learning
easier to implement. We can identify compact
nogoods with different parts of the decomposition.
Third, the variables introduced in such decompositions
give ths solver access to the state
of the propagator. This may be useful
when making branching decisions. 
Fourth, such decompositions can often
be encoded effectively into SAT and linear inequalities. We can
thereby provide the power of global propagation algorithms
to SAT and IP solvers. Finally, we expect many
other global constraints that count variables and values to be 
decomposable in similar ways. For instance,
it is known that BC on the \softalldiff constraint
is equivalent to BC on the \AtLeastNValue constraint 
\cite{bhhkwconstraint2006}.
We are currently studying decompositions
of other global constraints, such as the
\usedby \cite{bktcpaior04}
and other soft global constraints such as 
\softgcc \cite{softgcc}.

\section*{Acknowledgements}

{NICTA is funded by 
the Australian Government's Department of Broadband, 
Communications and the Digital Economy, and the 
Australian Research Council through ``Backing Australia's Ability''
and the ICT Centre of Excellence programmes.
Christian Bessiere is supported by the ANR project  ANR-06-BLAN-0383-02.}

\bibliographystyle{splncs}

\bibliography{/Users/twalsh/Documents/biblio/a-z,/Users/twalsh/Documents/biblio/pub,/Users/twalsh/Documents/biblio/a-z2,/Users/twalsh/Documents/biblio/pub2}


\end{document}